\documentclass{article}
\usepackage{spconf,amsmath,graphicx}
\usepackage{amsfonts}
\usepackage{amssymb}
\usepackage{url}
\usepackage{float}
\title{Attention-based Wav2Text with Feature Transfer Learning}
\name{Andros Tjandra, Sakriani Sakti, Satoshi Nakamura}
\address{Graduate School of Information Science, Nara Institute of Science and Technology, Japan \\
}

\begin{document}
	
	\maketitle
	\begin{abstract}
Conventional automatic speech recognition (ASR) typically performs multi-level pattern recognition tasks that map the acoustic speech waveform into a hierarchy of speech units. But, it is widely known that information loss in the earlier stage can propagate through the later stages. After the resurgence of deep learning, interest has emerged in the possibility of developing a purely end-to-end ASR system from the raw waveform to the transcription without any predefined alignments and hand-engineered models. However, the successful attempts in end-to-end architecture still used spectral-based features, while the successful attempts in using raw waveform were still based on the hybrid deep neural network - Hidden Markov model (DNN-HMM) framework. In this paper, we construct the first end-to-end attention-based encoder-decoder model to process directly from raw speech waveform to the text transcription. We called the model as “Attention-based Wav2Text”. To assist the training process of the end-to-end model, we propose to utilize a feature transfer learning. Experimental results also reveal that the proposed Attention-based Wav2Text model directly with raw waveform could achieve a better result in comparison with the attentional encoder-decoder model trained on standard front-end filterbank features.
	\end{abstract}
	\noindent\textbf{Index Terms}: speech recognition, end-to-end neural network, raw speech waveform
	
	\section{Introduction}
	Conventional large-vocabulary continuous speech recognition (LVCSR) systems typically perform multi-level pattern recognition tasks that map the acoustic speech waveform into a hierarchy of speech units such as sub-words (phonemes), words, and strings of words (sentences). Such systems basically consist of several sub-components (feature extractor, acoustic model, pronunciation lexicon, language model) that are trained and tuned separately \cite{gales2008application}. First, the speech signal is processed into a set of observation features based on a carefully hand-crafted feature extractor, such as Mel frequency cepstral coefficients (MFCC) or Mel-scale spectrogram. Then the acoustic model classifies the observation features into sub-unit or phoneme classes. Finally, the search algorithm finds the most probable word sequence based on the evidence of the acoustic model, the lexicon, and the language model. But, it is widely known that information loss in the earlier stage can propagate through the later stages.
	
	Deep learning algorithms have produced many state-of-the-art performances in various tasks that have revitalized the use of neural networks for ASR. One of the important factors behind the popularity of deep learning is the possibility of simplifying many complicated hand-engineered models by letting DNNs find their way to map from input to output spaces. Interest has emerged recently in the possibility of learning DNN-based acoustic models directly from the raw speech waveform without any predefined alignments and hand-engineered models. In this way, the feature extractor and acoustic model can be integrated into a single architecture. Palaz et al. \cite{palaz2013end, palaz2015convolutional} proposed a convolutional neural network (CNN) to directly train an acoustic model from the raw speech waveform. Sainath et al. \cite{sainath2015learning} used time-convolutional layers over raw speech and trained them jointly with the long short-term memory deep neural network (CLDNN) acoustic model. The results showed that raw waveform CLDNNs matched the performance of log-mel CLDNNs on a voice search task. Ghahremani et al. \cite{ghahremani2016acoustic} recently proposed a CNN time-delay neural network (CNN-TDNN) with network-in-network (NIN) architecture, and also showed that their model outperformed MFCC-based TDNN on the Wall Street Journal (WSJ) \cite{paul92wsj} task. But despite significant progress that has been made, the successful models were mostly demonstrated only within the hybrid DNN-HMM speech recognition frameworks.
	
	On the other hand, some existing works constructed end-to-end neural network models for ASR and replaced the acoustic model, the lexicon model, and the language model with a single integrated model, thus simplifying the pipeline. Graves et al. \cite{graves2006connectionist, graves2013speech} successfully built an end-to-end ASR based on the connectionist temporal classification (CTC) framework. Amodei et al. \cite{amodei2016deep} also constructed an end-to-end CTC-based ASR that directly produced character strings instead of phoneme sequences. But the CTC-based architecture still predicts the target outputs for every frame without any implicit knowledge about the language model. Another approach uses a sequence-to-sequence attention-based encoder-decoder that explicitly uses the history of previous outputs. Chorowski et al. \cite{chorowski2014end} and Chan et al.\cite{chan2016listen} has successfully demonstrated encoder-decoder based ASR frameworks. Unfortunately, most of these works still used the standard spectral features (i.e., Mel-scale spectrogram, MFCC) as the input. The only attempt on end-to-end speech recognition for a raw waveform was recently proposed by \cite{collobert2016wav2letter}. Their system used a deep CNN and was trained with the automatic segmentation criterion (ASG) as an alternative to CTC. However, similar with CTC, the model did not explicitly use the history of the previous outputs assuming they were conditionally independent of each other. Furthermore, its performance was only reported using a very large data set (about 1000h of audio files).
	
	To the best of our knowledge, few studies have explored a single end-to-end ASR architecture  trained on raw speech waveforms to directly output text transcription, and none of those models were built based on an encoder-decoder architecture. In this paper, we take a step forward to construct an end-to-end ASR using an attentional-based encoder-decoder model for processing raw speech waveform, naming it as ``Attention-based Wav2Text". We investigate the performance of our proposed models on standard ASR datasets. In practice, optimizing an encoder-decoder framework is more difficult than a standard neural network architecture \cite{chan2016listen}. Therefore, we propose a feature transfer learning method to assist the training process for our end-to-end attention-based ASR model.
	
	\section{Attention-based Encoder Decoder for Raw Speech Recognition}

	The encoder-decoder model is a neural network that directly models conditional probability $p(\mathbf{y}|\mathbf{x})$ where $\mathbf{x} = [x_1, ..., x_S]$ is the source sequence with length $S$ and $\mathbf{y} = [y_1, ..., y_T]$ is the target sequence with length $T$. It consists of  encoder, decoder and attention modules. The encoder task processes an input sequence $\mathbf{x}$ and outputs representative information $\mathbf{h^e} = [h^e_1, ...,h^e_S]$ for the decoder. The attention module is an extension scheme that assists the decoder to find relevant information on the encoder side based on the current decoder hidden states \cite{bahdanau2014neural, luong2015effective}. Usually, the attention module produces context information $c_t$ at time $t$ based on the encoder and decoder hidden states:
	\begin{align}
		c_t &= \sum_{s=1}^{S} a_t(s) * h^e_s \\
		a_t(s) &= \text{Align}({h^e_s}, h^d_t) = \frac{\exp(\text{Score}(h^e_s, h^d_t))}{\sum_{s=1}^{S}\exp(\text{Score}(h^e_s, h^d_t))}.
	\end{align}
	There are several variations for score function  :
	\begin{align}
		\text{Score}(h_s^e, h_t^d) =
		\begin{cases}
			\langle h_s^e, h_t^d\rangle, & \text{dot product}  \\
			h_s^{e\intercal} W_{s} h_t^d, & \text{bilinear}  \\
			V_s^{\intercal} \tanh(W_{s} [h_s^e, h_t^d]), & \text{MLP} \label{eq:mlpscore}  \\
		\end{cases}
	\end{align} where $\text{Score}:(\mathbb{R}^M \times \mathbb{R}^N) \rightarrow \mathbb{R}$, $M$ is the number of hidden units for the encoder and $N$ is the number of hidden units for the decoder.
	Finally, the decoder task, which predicts the target sequence probability at time $t$ based on previous output and context information $c_t$, can be formulated as:

	\begin{equation}
		\log{p(\mathbf{y}|\mathbf{x})} = \sum_{t=1}^{T}\log{p(y_t|y_{<t}, c_t)}
	\end{equation}
	The most common input $\mathbf{x}$ for speech recognition tasks is a sequence of feature vectors such as log Mel-spectral spectrogram and/or MFCC. Therefore, $ \mathbf{x} \in \mathbb{R}^{S \times D}$ where D is the number of the features and S is the total length of the utterance in frames. The output $\mathbf{y}$ can be either phoneme or grapheme (character) sequence.
	
\begin{figure}[t]
		\centering
		\includegraphics[width=0.9\linewidth]{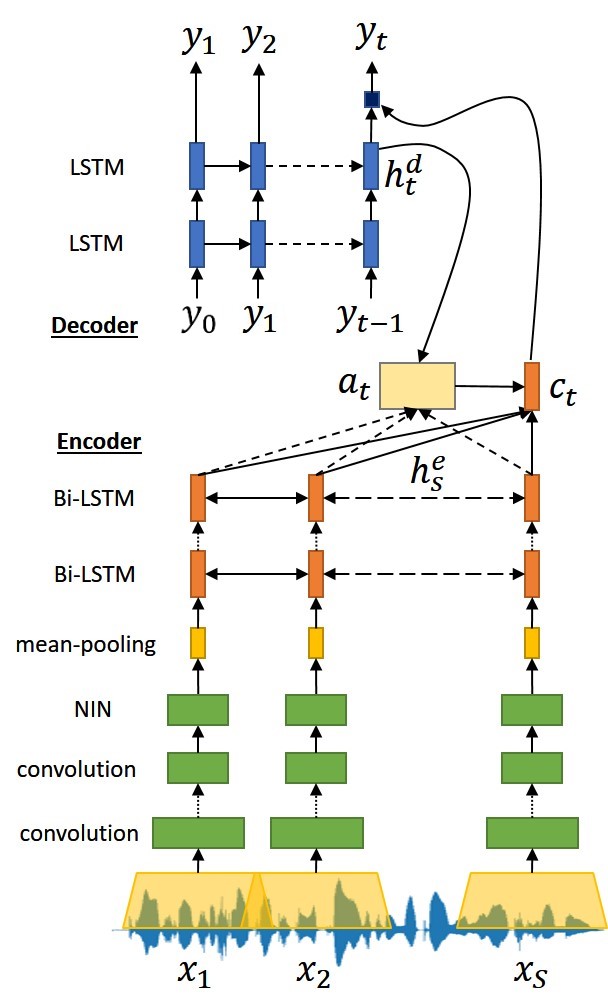}
		\caption{Attention-based Wav2Text architecture.}
		\label{fig:rawatte2e}
	\end{figure}

	In this work, we use the raw waveform as the input representation instead of spectral-based features and a grapheme (character) sequence as the output representation. In contrast to most encoder-decoder architectures, which are purely based on recurrent neural network (RNNs) framework, we construct an encoder with several convolutional layers \cite{lecun1989backpropagation} followed by NIN layers \cite{lin2013network} as the lower part in the encoder and integrate them with deep bidirectional long short-term memory (Bi-LSTM) \cite{hochreiter1997long} at the higher part. We use convolutional layers because they are suitable for extracting local information from raw speech. We use a striding mechanism to reduce the dimension from the input frames \cite{springenberg2014striving}, while the NIN layer represents more complex structures on the top of the convolutional layers. On the decoder side, we use a standard deep unidirectional LSTM with global attention \cite{luong2015effective} that is calculated by a multi-layer perceptron (MLP) as described in Eq. \ref{eq:mlpscore}. For more details, we illustrate our architecture in Figure~\ref{fig:rawatte2e}.

	\section{Feature Transfer Learning}

	Deep learning is well known for its ability to learn directly from low-level feature representation such as raw speech \cite{palaz2013end, sainath2015learning}. However, in most cases such models are already conditioned on a fixed input size and a single target output (i.e., predicting one phoneme class for each input frame). In the attention-based encoder-decoder model, the training process is not as easy as in a standard neural network model \cite{chan2016listen} because the attention-based model needs to jointly optimize three different modules simultaneously: (1) an encoder module for producing representative information from a source sequence; (2) an attention module for calculating the correct alignment; and (3) a decoder module for generating correct transcriptions. If one of these modules has difficulty fulfilling its own tasks, then the model will fail to produce good results.

	\begin{figure}[h]
		\centering
		\includegraphics[width=0.9\linewidth]{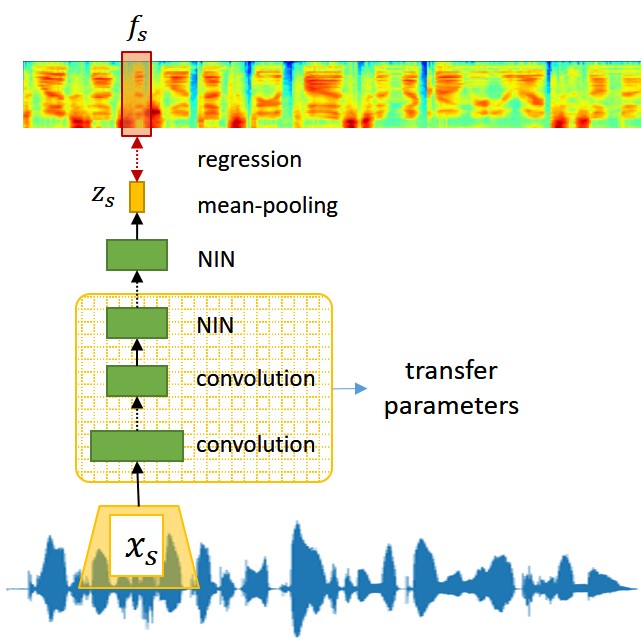}

		\caption{Feature transfer learning: train lower layers of the encoder (convolutional and NIN layers) to predict spectral features given corresponding raw waveform; then transfer the trained layers and parameters (marked by orange square) into attention-based encoder decoder model (see Figure~\ref{fig:rawatte2e}).}
		\label{fig:transfer_fbank}
	\end{figure}
	
	To ease the burden on training the whole encoder-decoder architecture directly to predict the text transcription given the raw speech waveform, we utilize a transfer learning method on the encoder part. Specifically, we only train the encoder's lower layers consisting of the convolutional and NIN layers to predict the spectral features given the corresponding raw waveform. In this work, we utilize two widely used spectral features: MFCC and log Mel-scale spectrogram as the transfer learning target. Figure~\ref{fig:transfer_fbank} shows our feature transfer learning architecture. First, given segmented raw speech waveform $\mathbf{x} = [x_1, ..., x_S] $, we extract corresponding $D$-dimensional spectral features $\mathbf{f} = [f_1,..,f_S], \quad \forall s, f_s \in \mathbb{R}^{D} $. Then we process raw speech $x_s$ with several convolutions, followed by NIN layers in the encoder part. In the last NIN-layer, we set a fixed number of channels as $D$ channels and apply mean-pooling across time. Finally, we get predictions for corresponding spectral features $z_s \in \mathbb{R}^{D}$ and optimize all of the parameters by minimizing the mean squared error between predicted spectral features $\mathbf{z}$ and target spectral features $\mathbf{f}$:

	\begin{equation} \label{eq:single_target}
		\mathcal{L}_{tf} = \frac{1}{S} \sum_{s=1}^{S} \sum_{d=1}^{D}(f_s(d) - z_s(d))^2.
	\end{equation}
	In this paper, we also explore multi target feature transfer using a similar structure as in Figure~\ref{fig:transfer_fbank} but with two parallel NIN layers, followed by mean-polling at the end. One of the output layers is used to predicts log Mel-scale spectrogram and another predicts MFCC features. We modify the single target loss function from Eq.~\ref{eq:single_target} into the following:

	\begin{align}
		\mathcal{L}_{tf} = \frac{1}{S} \sum_{s=1}^{S} &\left( \sum_{d=1}^{D_a}(f^a_s(d) - z^a_s(d))^2 +  \sum_{d=1}^{D_b}(f^b_s(d) - z^b_s(d))^2 \right).
	\end{align} where $z^a_s, z^b_s$ are the predicted Mel-scale spectrogram and the MFCC values, and $f^a_s, f^b_s$ are the real Mel-scale spectrogram and MFCC features for frame $s$.
	After optimizing all the convolutional and NIN layer parameters, we transfer the trained layers and parameters and integrate them with the Bi-LSTM encoder. Finally, we jointly optimize the whole structure together.
	
  \begin{figure*}[t]
		\centering
		\includegraphics[width=1.0\linewidth,height=6cm]{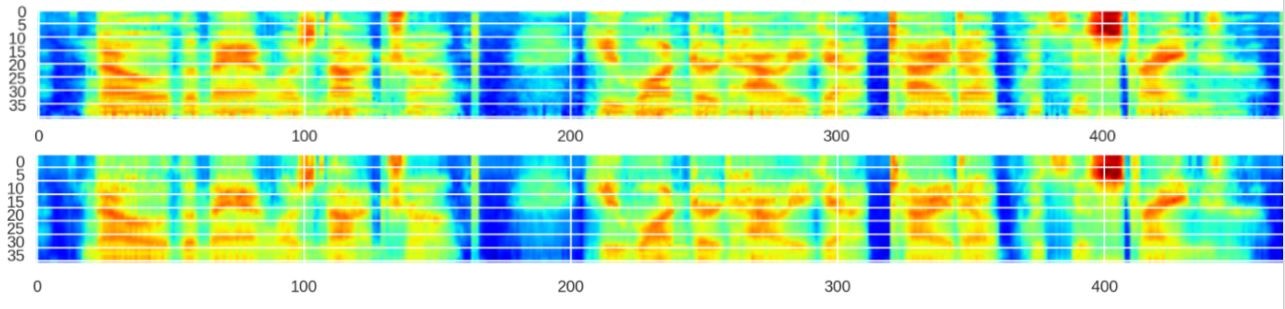}
		\caption{Example of our transfer learning model output: top is the original Mel-spectrogram, and bottom is the predicted Mel-spectrogram.}
		\label{fig:fbank_ori_pred}
	\end{figure*}

	\section{Experimental Setup and Results}

	\subsection{Speech Data}

	In this study, we investigate the performance of our proposed models on WSJ \cite{paul92wsj}. We used the same definitions of the training, development and test set as the Kaldi s5 recipe \cite{povey11asru}. The raw speech waveforms were segmented into multiple frames with a 25ms window size and a 10ms step size. We normalized the raw speech waveform into the range -1 to 1. For spectral based features such as MFCC and log Mel-spectrogram, we normalized the features for each dimension into zero mean and unit variance. For WSJ, we separated into two experiments by using WSJ-SI84 only and WSJ-SI284 data. We used dev\_93 for our validation set and eval\_92 for our test set. We used the character sequence as our decoder target and followed the preprocessing step proposed by \cite{hannun2014first}. The text from all the utterances was mapped into a 32-character set: 26 (a-z) alphabet, apostrophe, period, dash, space, noise, and ``eos".

	\subsection{Model Architectures}
	
	Our attention-based Wav2Text architecture uses four convolutional layers, followed by two NIN layers at the lower part of the encoder module. For all the convolutional layers, we used a leaky rectifier unit (LReLU)\cite{maas2013rectifier} activation function with leakiness $(l = 0.1)$. Inside the first NIN layers, we stacked three consecutive filters with LReLU activation function. For the second NIN layers, we stacked two consecutive filters with tanh and identity activation function. For the feature transfer learning training phase, we used Momentum SGD with a learning rate of 0.01 and momentum of 0.9. Table~\ref{enc-setting} summarizes the details of the layer settings for the convolutional and NIN layers.
	\begin{table}[h]
		\centering
		%\small
		\caption{Layer setting details for convolutional and NIN layers. Sorted from the input layer to the output layer.}
        \vspace{0.2cm}
		\label{enc-setting}
		\begin{tabular}{|c|c|c|c|c|}
			\hline
			\textbf{\begin{tabular}[c]{@{}c@{}}Layer\\ (Transfer)\end{tabular}} & \textbf{Channels}                                              & \textbf{Filter} & \textbf{Stride} & \textbf{Act.}  \\ \hline
			1D Conv (\checkmark)                                                & 128                                                            & 80                                                              & 4               & LReLU          \\ \hline
			1D Conv (\checkmark)                                                & 128                                                            & 25                                                               & 2               & LReLU          \\ \hline
			1D Conv (\checkmark)                                                & 128                                                            & 10                                                               & 1               & LReLU          \\ \hline
			1D Conv (\checkmark)                                                & 128                                                            & 5                                                                & 1               & LReLU          \\ \hline
			NIN (\checkmark)                                                    & [128,128] & 1                                                                & 1               & LReLU$\times$2 \\ \hline
			NIN ($\times$)                                                      & {[}128,N{]}                                                    & 1                                                                & 1               & \begin{tabular}[c]{@{}l@{}}Tanh \\ Identity\end{tabular}    \\ \hline
		\end{tabular}
	\end{table}
	
	On the top layers of the encoder after the transferred convolutional and NIN layers, we put three bidirectional LSTMs (Bi-LSTM) with 256 hidden units (total 512 units for both directions). To reduce the computational time, we used hierarchical subsampling \cite{graves2012supervised, bahdanau2016end, chan2016listen}. We applied subsampling on all the Bi-LSTM layers and reduced the length by a factor of 8.
	
	On the decoder side, the previous input phonemes / characters were converted into real vectors by a 128-dimensional embedding matrix. We used one unidirectional LSTM with 512 hidden units and followed by a softmax layer to output the character probability. For the end-to-end training phase, we froze the parameter values from the transferred layers from epoch 0 to epoch 10, and after epoch 10 we jointly optimized all the parameters together until the end of training (a total 40 epochs). We used an Adam \cite{kingma2014adam} optimizer with a learning rate of 0.0005.
	
	In the decoding phase, we used a beam search strategy with beam size $= 5$ and we adjusted the score by dividing with the transcription length to prevent the decoder from favoring shorter transcriptions. We did not use any language model or lexicon dictionary for decoding. All of our models were implemented on the PyTorch framework \footnote{PyTorch \url{https://github.com/pytorch/pytorch}}.
	
	For comparison, we also evaluated the standard attention-based encoder decoder with Mel-scale spectrogram input as the baseline. Here, we used similar settings as the proposed model, except we replaced the convolutional and NIN layers with a feedforward layer (512 hidden units).
	
	\subsection{Result}
	\begin{table}[H]
		\centering
		%\small
		\caption{Character error rate (CER) result from baseline and proposed models on WSJ0 and WSJ1 dataset. All of these results are produced without using language model or lexicon dictionary. Word error rate (WER) for Att Wav2Text + transfer multi-target is 17.04\%, compared to Joint CTC+Att (MTL)\cite{kim2016joint} 18.2\% and standard Enc-Dec Att \cite{bahdanau2016end} 18.6\%.}
		\vspace{0.2cm}
		\label{tbl:all}
		\begin{tabular}{|l|l|l|}
			\hline
			\multicolumn{1}{|c|}{\textbf{Models}}
			& \multicolumn{1}{c|}{\textbf{Features}}
			& \multicolumn{1}{c|}{\textbf{Results}} \\ \hline \hline
			\multicolumn{2}{|c|}{\textbf{WSJ-SI84}}
			& \multicolumn{1}{c|}{\textbf{CER (\%)}}          \\ \hline
			\multicolumn{3}{|c|}{Baseline}                                                                                                                                               \\ \hline
			CTC \cite{kim2016joint}     & fbank                & 20.34\%                                       \\ \hline
			Att Enc-Dec Content \cite{kim2016joint}  & fbank                & 20.06\%                                       \\ \hline
			Att Enc-Dec Location \cite{kim2016joint}  & fbank                & 17.01\%                                       \\ \hline
			Joint CTC+Att (MTL) \cite{kim2016joint} & fbank                & 14.53\%                 \\ \hline
			Att Enc-Dec (ours) & fbank                & 17.68\%                                             \\ \hline
			\multicolumn{3}{|c|}{Proposed}                                                                                                                                               \\ \hline
			\begin{tabular}[c]{@{}l@{}}Att Wav2Text \\(direct)\end{tabular}        & raw speech
			& \begin{tabular}[c]{@{}l@{}} (not\\converged)\end{tabular}
			\\ \hline
			\begin{tabular}[c]{@{}l@{}}Att Wav2Text \\(transfer from fbank)\end{tabular}        & raw speech  &  16.87 \%                                     \\ \hline
			\begin{tabular}[c]{@{}l@{}}Att Wav2Text \\ (transfer from MFCC)\end{tabular}         & raw speech  & 15.74\%                                       \\ \hline
			\begin{tabular}[c]{@{}l@{}}Att Wav2Text \\ (transfer from multi target)\end{tabular} & raw speech  & 14.71\%                                        \\ \hline
			\hline		
			\multicolumn{2}{|c|}{\textbf{WSJ-SI284}}
			& \multicolumn{1}{c|}{\textbf{CER (\%)}}          \\ \hline
			\multicolumn{3}{|c|}{Baseline} \\ \hline
			CTC \cite{kim2016joint}     & fbank                & 8.97\%                                       \\ \hline
			Att Enc-Dec Content\cite{kim2016joint}  & fbank                & 11.08\%                                       \\ \hline
			Att Enc-Dec Location\cite{kim2016joint}  & fbank                & 8.17\%                                       \\ \hline
			Joint CTC+Att (MTL) \cite{kim2016joint} & fbank                & 7.36\%                 \\ \hline
			Att Enc-Dec (ours)  & fbank                & 7.69\%                                       \\ \hline
			\multicolumn{3}{|c|}{Proposed}                                                                                                                                               \\ \hline
			\begin{tabular}[c]{@{}l@{}}Att Wav2Text\\ (direct)\end{tabular}        & raw speech
			& \begin{tabular}[c]{@{}l@{}} (not\\converged)\end{tabular}
			\\ \hline
			\begin{tabular}[c]{@{}l@{}}Att Wav2Text \\ (transfer from fbank)\end{tabular}        & raw speech  &  6.78 \%                                     \\ \hline
			\begin{tabular}[c]{@{}l@{}}Att Wav2Text \\(transfer from MFCC)\end{tabular}         & raw speech  & 6.58\%                                       \\ \hline
			\begin{tabular}[c]{@{}l@{}}Att Wav2Text\\
				(transfer from multi target)\end{tabular} & raw speech  & 6.54\%                                        \\ \hline
		\end{tabular}
	\end{table}
An example of our transfer learning results is shown in Figure~\ref{fig:fbank_ori_pred}, and Table \ref{tbl:all} shows the speech recognition performance in CER for both the WSJ-SI84 and WSJ-SI284 datasets. 
We compared our method with several published models like CTC, Attention Encoder-Decoder and Joint CTC-Attention model that utilize CTC for training the encoder part. Besides, we also train our own baseline Attention Encoder-Decoder with Mel-scale spectrogram. The difference between our Attention Encoder-Decoder (``Att Enc-Dec (ours)", ``Att Enc-Dec Wav2Text") with Attention Encoder-Decoder from \cite{kim2016joint} (``Att Enc-Dec Content", ``Att Enc-Dec Location") is we used the current hidden states to generate the attention vector instead of the previous hidden states. Another addition is we utilized ``input feedback" method \cite{luong2015effective} by concatenating the previous context vector into the current input along with the character embedding vector. By using those modifications, we are able to improve the baseline performance.

Our proposed Wav2Text models without any transfer learning failed to converge. In contrast, with transfer learning, they significantly surpassed the performance of the CTC and encoder-decoder from Mel-scale spectrogram features. This suggests that by using transfer learning for initializing the lower part of the encoder parameters, our model also performed better then their original features.

\section{Related Work}
	Transfer learning is the ability of a learning algorithm to convey knowledge across different tasks. The initial idea is to reuse previously obtained knowledge to enhance the learning for new things. The standard procedure are : first, train the model on a base dataset and task, then the learned features and/or parameters are reused for learning a second target dataset and task. Bengio et al. \cite{bengio2013representation} provided deep reviews about multi-task and transfer learning on deep learning models. Jason et al. \cite{yosinski2014transferable} showed that a model with transferred parameter consistently outperformed a randomly initialized one.
	
	In speech recognition research, transfer learning has been studied for many years, including successful cases of speaker adaptation and cross-lingual acoustic modeling \cite{das2015cross}. One popular scheme for utilizing DNNs for transfer learning within ASR frameworks is a tandem approach \cite{grezl2007probabilistic}.
	This idea first trains a DNN with a narrow hidden bottleneck layer to perform phoneme classification at the frame level and then reuses the activations from the narrow hidden bottleneck layer as discriminative features in conventional GMM-HMM or hybrid DNN-HMM models \cite{YAN}. Another study introduced a convolutional bottleneck network as an alternative tandem bottleneck feature architecture \cite{VAS}. However, although such a feature transfer learning framework provides many advantages in ASR, the usage in an end-to-end attention-based ASR framework has not been explored.
	
	This study performs feature transfer learning on the encoder part of the end-to-end attention-based ASR architecture. We train the convolutional encoder to predict the spectral features given the corresponding raw speech waveform. After that, we transfer the trained layers and parameters, integrate them with the LSTM encoder-decoder, and eventually optimize the whole structure to predict the correct output text transcription given the raw speech waveform.

	\section{Conclusion}

	This paper described the first attempt to build an end-to-end attention-based encoder-decoder speech recognition that directly predicts the text transcription given raw speech input. We also proposed feature transfer learning to assist the encoder-decoder model training process and presented a novel architecture that combined convolutional, NIN and Bi-LSTM layers into a single encoder part for raw speech recognition. Our results suggest that transfer learning is a very helpful method for constructing an end-to-end system from such low-level features as raw speech signals. With transferred parameters, our proposed attention-based Wav2Text models converged and matched the performance with the attention-based encoder-decoder model trained on standard spectral-based features. The best performance was achieved by Wav2Text models with transfer learning from multi target scheme.

	\section{Acknowledgements}

Part of this work was supported by JSPS KAKENHI Grant Numbers JP17H06101 and JP 17K00237.
	\bibliographystyle{IEEEbib}
	\bibliography{strings,refs}

\begin{thebibliography}{10}

\bibitem{gales2008application}
Mark Gales and Steve Young,
\newblock ``The application of hidden {M}arkov models in speech recognition,''
\newblock {\em Foundations and {T}rends in {S}ignal {P}rocessing}, vol. 1, no.
  3, pp. 195--304, 2008.

\bibitem{palaz2013end}
Dimitri Palaz, Ronan Collobert, and Mathew~Magimai Doss,
\newblock ``End-to-end phoneme sequence recognition using convolutional neural
  networks,''
\newblock {\em arXiv preprint arXiv:1312.2137}, 2013.

\bibitem{palaz2015convolutional}
Dimitri Palaz, Mathew~Magimai Doss, and Ronan Collobert,
\newblock ``Convolutional neural networks-based continuous speech recognition
  using raw speech signal,''
\newblock in {\em Acoustics, Speech and Signal Processing (ICASSP), 2015 IEEE
  International Conference on}. IEEE, 2015, pp. 4295--4299.

\bibitem{sainath2015learning}
Tara~N Sainath, Ron~J Weiss, Andrew~W Senior, Kevin~W Wilson, and Oriol
  Vinyals,
\newblock ``Learning the speech front-end with raw waveform {CLDNN}s.,''
\newblock in {\em Interspeech}, 2015, vol. 2015.

\bibitem{ghahremani2016acoustic}
Pegah Ghahremani, Vimal Manohar, Daniel Povey, and Sanjeev Khudanpur,
\newblock ``Acoustic modelling from the signal domain using {CNNs},''
\newblock in {\em Interspeech}, 2016, vol. 2016.

\bibitem{paul92wsj}
Douglas~B. Paul and Janet~M. Baker,
\newblock ``The design for the {W}all {S}treet {J}ournal-based {CSR} corpus,''
\newblock in {\em Proceedings of the Workshop on Speech and Natural Language},
  Stroudsburg, PA, USA, 1992, HLT '91, pp. 357--362, Association for
  Computational Linguistics.

\bibitem{graves2006connectionist}
Alex Graves, Santiago Fern{\'a}ndez, Faustino Gomez, and J{\"u}rgen
  Schmidhuber,
\newblock ``Connectionist temporal classification: labelling unsegmented
  sequence data with recurrent neural networks,''
\newblock in {\em Proceedings of the 23rd International Conference on Machine
  learning}. ACM, 2006, pp. 369--376.

\bibitem{graves2013speech}
Alex Graves, Abdel~Rahman Mohamed, and Geoffrey Hinton,
\newblock ``Speech recognition with deep recurrent neural networks,''
\newblock in {\em Acoustics, Speech and Signal processing (ICASSP), 2013 IEEE
  International Conference on}. IEEE, 2013, pp. 6645--6649.

\bibitem{amodei2016deep}
Dario Amodei, Rishita Anubhai, Eric Battenberg, Carl Case, Jared Casper, Bryan
  Catanzaro, JingDong Chen, Mike Chrzanowski, Adam Coates, Greg Diamos, et~al.,
\newblock ``Deep speech 2: End-to-end speech recognition in {E}nglish and
  {M}andarin,''
\newblock in {\em Proceedings of The 33rd International Conference on Machine
  Learning}, 2016, pp. 173--182.

\bibitem{chorowski2014end}
Jan Chorowski, Dzmitry Bahdanau, Kyunghyun Cho, and Yoshua Bengio,
\newblock ``End-to-end continuous speech recognition using attention-based
  recurrent {NN}: First results,''
\newblock {\em arXiv preprint arXiv:1412.1602}, 2014.

\bibitem{chan2016listen}
William Chan, Navdeep Jaitly, Quoc Le, and Oriol Vinyals,
\newblock ``Listen, attend and spell: A neural network for large vocabulary
  conversational speech recognition,''
\newblock in {\em Acoustics, Speech and Signal Processing (ICASSP), 2016 IEEE
  International Conference on}. IEEE, 2016, pp. 4960--4964.

\bibitem{collobert2016wav2letter}
Ronan Collobert, Christian Puhrsch, and Gabriel Synnaeve,
\newblock ``Wav2letter: an end-to-end convnet-based speech recognition
  system,''
\newblock {\em arXiv preprint arXiv:1609.03193}, 2016.

\bibitem{bahdanau2014neural}
Dzmitry Bahdanau, Kyunghyun Cho, and Yoshua Bengio,
\newblock ``Neural machine translation by jointly learning to align and
  translate,''
\newblock {\em arXiv preprint arXiv:1409.0473}, 2014.

\bibitem{luong2015effective}
Minh-Thang Luong, Hieu Pham, and Christopher~D Manning,
\newblock ``Effective approaches to attention-based neural machine
  translation,''
\newblock {\em arXiv preprint arXiv:1508.04025}, 2015.

\bibitem{lecun1989backpropagation}
Yann LeCun, Bernhard Boser, John~S Denker, Donnie Henderson, Richard~E Howard,
  Wayne Hubbard, and Lawrence~D Jackel,
\newblock ``Backpropagation applied to handwritten zip code recognition,''
\newblock {\em Neural computation}, vol. 1, no. 4, pp. 541--551, 1989.

\bibitem{lin2013network}
Min Lin, Qiang Chen, and Shuicheng Yan,
\newblock ``Network in network,''
\newblock {\em arXiv preprint arXiv:1312.4400}, 2013.

\bibitem{hochreiter1997long}
Sepp Hochreiter and J{\"u}rgen Schmidhuber,
\newblock ``Long short-term memory,''
\newblock {\em Neural computation}, vol. 9, no. 8, pp. 1735--1780, 1997.

\bibitem{springenberg2014striving}
Jost~Tobias Springenberg, Alexey Dosovitskiy, Thomas Brox, and Martin
  Riedmiller,
\newblock ``Striving for simplicity: The all convolutional net,''
\newblock {\em arXiv preprint arXiv:1412.6806}, 2014.

\bibitem{povey11asru}
Daniel Povey, Arnab Ghoshal, Gilles Boulianne, Lukas Burget, Ondrej Glembek,
  Nagendra Goel, Mirko Hannemann, Petr Motlicek, Yanmin Qian, Petr Schwarz, Jan
  Silovsky, Georg Stemmer, and Karel Vesely,
\newblock ``The {Kaldi} speech recognition toolkit,''
\newblock in {\em IEEE 2011 Workshop on Automatic Speech Recognition and
  Understanding}. Dec. 2011, IEEE Signal Processing Society,
\newblock IEEE Catalog No.: CFP11SRW-USB.

\bibitem{hannun2014first}
Awni~Y Hannun, Andrew~L Maas, Daniel Jurafsky, and Andrew~Y Ng,
\newblock ``First-pass large vocabulary continuous speech recognition using
  bi-directional recurrent {DNN}s,''
\newblock {\em arXiv preprint arXiv:1408.2873}, 2014.

\bibitem{maas2013rectifier}
Andrew~L Maas, Awni~Y Hannun, and Andrew~Y Ng,
\newblock ``Rectifier nonlinearities improve neural network acoustic models,''
\newblock in {\em in ICML Workshop on Deep Learning for Audio, Speech and
  Language Processing}, 2013.

\bibitem{graves2012supervised}
Alex Graves,
\newblock ``Supervised sequence labelling,''
\newblock in {\em Supervised Sequence Labelling with Recurrent Neural
  Networks}, pp. 5--13. Springer, 2012.

\bibitem{bahdanau2016end}
Dzmitry Bahdanau, Jan Chorowski, Dmitriy Serdyuk, Philemon Brakel, and Yoshua
  Bengio,
\newblock ``End-to-end attention-based large vocabulary speech recognition,''
\newblock in {\em Acoustics, Speech and Signal Processing (ICASSP), 2016 IEEE
  International Conference on}. IEEE, 2016, pp. 4945--4949.

\bibitem{kingma2014adam}
Diederik Kingma and Jimmy Ba,
\newblock ``Adam: A method for stochastic optimization,''
\newblock {\em arXiv preprint arXiv:1412.6980}, 2014.

\bibitem{kim2016joint}
Suyoun Kim, Takaaki Hori, and Shinji Watanabe,
\newblock ``Joint {CTC}-attention based end-to-end speech recognition using
  multi-task learning,''
\newblock in {\em Acoustics, Speech and Signal processing (ICASSP), 2017 IEEE
  International Conference on}. IEEE, 2017, p. to appear.

\bibitem{bengio2013representation}
Yoshua Bengio, Aaron Courville, and Pascal Vincent,
\newblock ``Representation learning: A review and new perspectives,''
\newblock {\em IEEE {T}ransactions on {P}attern {A}nalysis and {M}achine
  {I}ntelligence}, vol. 35, no. 8, pp. 1798--1828, 2013.

\bibitem{yosinski2014transferable}
Jason Yosinski, Jeff Clune, Yoshua Bengio, and Hod Lipson,
\newblock ``How transferable are features in deep neural networks?,''
\newblock in {\em Advances in neural information processing systems}, 2014, pp.
  3320--3328.

\bibitem{das2015cross}
Amit Das and Mark Hasegawa-Johnson,
\newblock ``Cross-lingual transfer learning during supervised training in low
  resource scenarios.,''
\newblock in {\em INTERSPEECH}, 2015, pp. 3531--3535.

\bibitem{grezl2007probabilistic}
Frantisek Gr{\'e}zl, Martin Karafi{\'a}t, Stanislav Kont{\'a}r, and Jan
  Cernocky,
\newblock ``Probabilistic and bottle-neck features for {LVCSR} of meetings,''
\newblock in {\em Acoustics, Speech and Signal Processing, 2007. ICASSP 2007.
  IEEE International Conference on}. IEEE, 2007, vol.~4, pp. IV--757.

\bibitem{YAN}
Z.J. Yan, Q.~Huo, and J.~Xu,
\newblock ``A scalable approach to using {DNN}-derived features in {GMM}-{HMM}
  based acoustic modeling for {LVCSR},''
\newblock in {\em Proc. INTERSPEECH}, 2013.

\bibitem{VAS}
K.~Vesely, M.~Karafi{\'a}t, and F.~Gr{\'e}zl,
\newblock ``Convolutive bottleneck network features for {LVCSR},''
\newblock in {\em Proc. ASRU}, Waikoloa, USA, 2011, pp. 42--47.

\end{thebibliography}

\end{document}